\documentclass[letterpaper, 10 pt, conference]{ieeeconf}

\IEEEoverridecommandlockouts
\usepackage[T1]{fontenc}
\usepackage[style=ieee, maxcitenames=2, mincitenames=1, natbib=true, dashed=false, maxbibnames=99, minbibnames=99]{biblatex}

\usepackage{amsmath}
\usepackage[caption=false]{subfig}

\let\labelindent\relax
\usepackage{enumitem}
\usepackage{tabularx, booktabs}

\usepackage{hyperref}
\usepackage{cleveref}
\usepackage{amsfonts}
\usepackage{graphicx}
\usepackage{balance} 
\usepackage{multirow}
\usepackage{siunitx}
\usepackage{newfloat}
\usepackage{mathtools}

\usepackage{setspace}
\Crefname{equation}{Eq.}{Eqs.}
\title{\LARGE \bf

Cooperative Informative Sensing for Monitoring Dynamic \\Indoor Environments via Multi-Agent Reinforcement Learning}

\author{
Kanghoon Lee$^\star$, Matthew M. Sato$^\star$, Jinnyeong Yang, Seungro Lee, Sujin Lee, \\
Jiachen Li, Kuk-Jin Yoon, Jinkyoo Park, Kincho H. Law, Yoonjin Yoon$^\dag$%
    \thanks{$^\star$Both authors contributed equally to this research.}%
    \thanks{$^\dag$Corresponding author.}%
    \thanks{K. Lee, J. Yang, S. Lee, K. Yoon, J. Park, and Y. Yoon are with the Korea Advanced Institute of Science and Technology (KAIST), Daejeon, Republic of Korea. {\tt\small \{leehoon, jinnyeong6118, roy.seungro.lee, kjyoon, jinkyoo.park, yoonjin\}@kaist.ac.kr}.}%
    \thanks{M. Sato and K. Law are with Stanford University, Stanford, CA, USA. {\tt\small \{satomm, law\}@stanford.edu}.}%
    \thanks{S. Lee is with the University of California, Berkeley, Berkeley, CA, USA. {\tt\small sujin.lee@berkeley.edu}.}
    \thanks{J. Li is with the University of California, Riverside (UCR), CA, USA. {\tt\small jiachen.li@ucr.edu}.}
}

\begin{document}

\maketitle

\begin{abstract}
Monitoring human activity in indoor environments is important for applications such as facility management, safety assessment, and space utilization analysis. While mobile robot teams offer the potential to actively improve observation quality, existing multi-robot monitoring and active perception approaches typically rely on coverage or visitation based objectives that are weakly aligned with the accuracy requirements of human-centric monitoring tasks. In this work, we formulate cooperative active observation as a decentralized control problem in which multiple robots adjust their motion to directly optimize monitoring accuracy under partial observability. We propose a learning-based framework for cooperative policies from decentralized observations using multi-agent reinforcement learning (MARL), supported by an architecture that handles variable numbers of humans and temporal dependencies. Simulation results across diverse indoor environments and monitoring tasks show that the proposed approach consistently outperforms classical coverage, persistent monitoring, and learning-free multi-robot baselines, while remaining robust to changes in the number of observed humans.
\end{abstract}

\section{Introduction}

Understanding how humans utilize indoor spaces is important for facility managers, developers, and engineers, enabling better building reconfiguration and informing the design of new buildings to improve efficiency and user experience \cite{hassanain2010analysis}. Accurate observations of human presence, movement, and interactions are essential for stakeholders, providing the data to estimate occupancy and activity patterns within the space. While many indoor environments already feature sensing infrastructure such as fixed cameras, indoor geometry (walls and corridors) imposes strong constraints on observation, leading to frequent occlusions. Because human motion is dynamic and complex \cite{rudenko2020human}, continuous observation from fixed viewpoints is challenging, resulting in sparse and fragmented observations that leave large portions of the environment unobserved for prolonged intervals. As a result, human-centric monitoring tasks that require accurate and continuous observations cannot be adequately addressed by static sensing systems or a single mobile robot.

To address sensing limitations, a variety of approaches have been studied in active perception \cite{bajcsy1988active}, informative path planning \cite{bourgault2002information}, multi-robot coverage \cite{chiun2025marvel}, and persistent monitoring \cite{smith2011persistent}. Classical informative path planning methods typically optimize trajectories to maximize information gain or coverage, often assuming static or slowly varying processes. Similarly, multi-robot coverage and persistent monitoring strategies emphasize uniform or frequent visitation over time, without explicitly accounting for the distribution of human behaviors. While effective for tasks such as environmental mapping or long-term surveillance, these approaches are less suited for monitoring dynamic human environments, where the relevance of observations evolves over time. Moreover, existing methods generally emphasize visitation or observability metrics, rather than directly optimizing for the accuracy of downstream human-centric estimates derived from partial observations, which is the focus of this work.

\begin{figure}[t!]
\centering
  \includegraphics[width=0.42\textwidth]{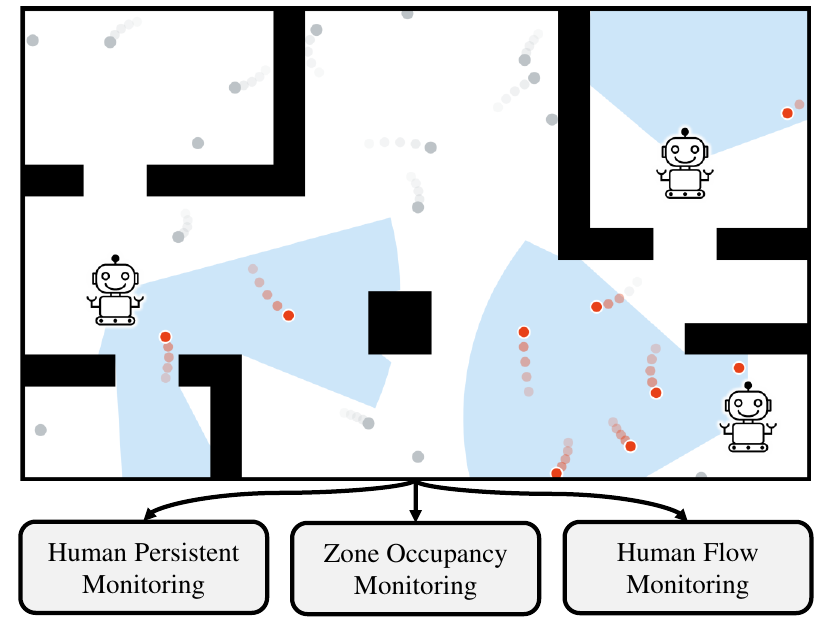}
  \vspace{-5pt}
\caption{\textbf{Illustration of cooperative active observation for human-centric monitoring in indoor environments.}
Red/gray dots denote visible/non-visible human agents, whose observability is limited by indoor occlusions and the robots’ restricted fields of view (FoV). Mobile robots actively control their motion to acquire observations within their FoV (shaded blue regions), supporting human persistent monitoring at the individual, zone, and inter-zone levels under partial observability.}
\vspace{-15pt}
\label{fig:teaser}
\end{figure}

In this work, we reframe active observation in indoor environments from traditional coverage or visitation objectives to a multi-agent decision-making framework using multiple mobile robots. Specifically, robots must determine where and when to observe to reduce uncertainty in task-relevant quantities derived from latent human dynamics, such as human-centric monitoring at the individual, zone, and inter-zone levels as shown in \Cref{fig:teaser}. We use cooperative sensing to improve the accuracy of downstream human activity estimates under partial observability. This formulation naturally captures the coupling between sensing actions and estimation quality, providing a unified framework for diverse human-centric monitoring tasks. 

To address the challenges of partial observability and dynamic human motion, we develop a learning-based framework for cooperative monitoring with multiple mobile robots. Since the problem involves decentralized observations and motion control, with success measured by the accuracy of human-centric monitoring, we adopt a multi-agent reinforcement learning (MARL) approach to learn cooperative policies in a closed-loop manner. Our policy architecture specifically supports variable-sized human observations and temporal dependencies, allowing robust operation regardless of the number of humans present. We evaluate the proposed approach through simulation studies and demonstrate consistent performance improvements over classical coverage, persistent monitoring, and learning-free multi-robot navigating approaches across a diverse set of monitoring tasks.

In summary, the main contributions are as follows:
\begin{itemize}[leftmargin=*]
    \item We formalize a cooperative informative sensing problem under partial observability that supports multiple monitoring tasks in dynamic indoor environments.
    \item We propose a learning-based multi-robot monitoring framework with a network architecture that effectively handles partial and variable-sized human observations, while enabling scalable inter-robot coordination.
    \item We validate the proposed approach by demonstrating improved performance over diverse multi-robot sensing methods across multiple monitoring tasks.
\end{itemize}

\section{Related Works}

\subsection{Active Observation and Information Gathering}
Active observation studies how an agent selects sensing actions such as viewpoints or trajectories to improve the quality of acquired information \cite{bajcsy1988active}. 
Classical robotics approaches include frontier-based exploration that move robots toward the boundary between known and unknown space \cite{yamauchi1997frontier}, and information-theoretic planning methods that maximize expected information gain \cite{bourgault2002information}.
In multi-robot settings, these principles have been extended to coordinated sensing and team-level information gathering under practical constraints such as limited communication or resources \cite{lauri2017multi,choudhury2020adaptive}.
Recently, deep RL and MARL have been applied to active observation for viewpoint control in target tracking \cite{jeong2021deep}, adaptive multi-agent tracking in dynamic environments \cite{ruckin2022adaptive}, and cooperative coverage and exploration \cite{chiun2025marvel}.
In our work, we address the cooperative sensing problem for dynamic indoor monitoring with multiple robots. Our approach optimizes auxiliary monitoring objectives derived from a shared belief state, enables transfer across monitoring tasks, and supports integration with existing fixed sensing infrastructure.

\begin{figure*}[t!]
\centering
\subfloat{
  \includegraphics[width=0.3\textwidth]{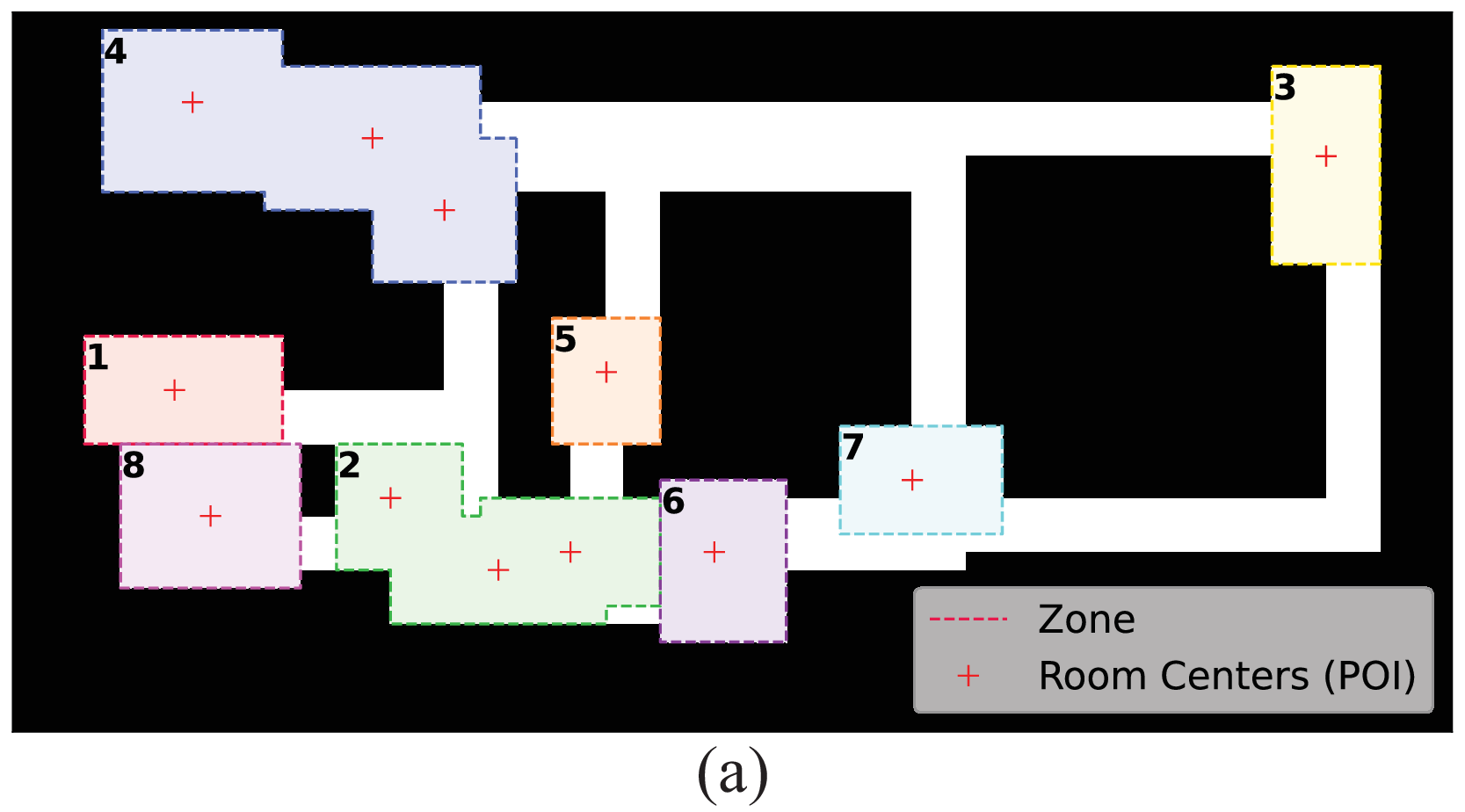}
  \label{fig:sim_a}
}
\subfloat{
  \includegraphics[width=0.3\textwidth]{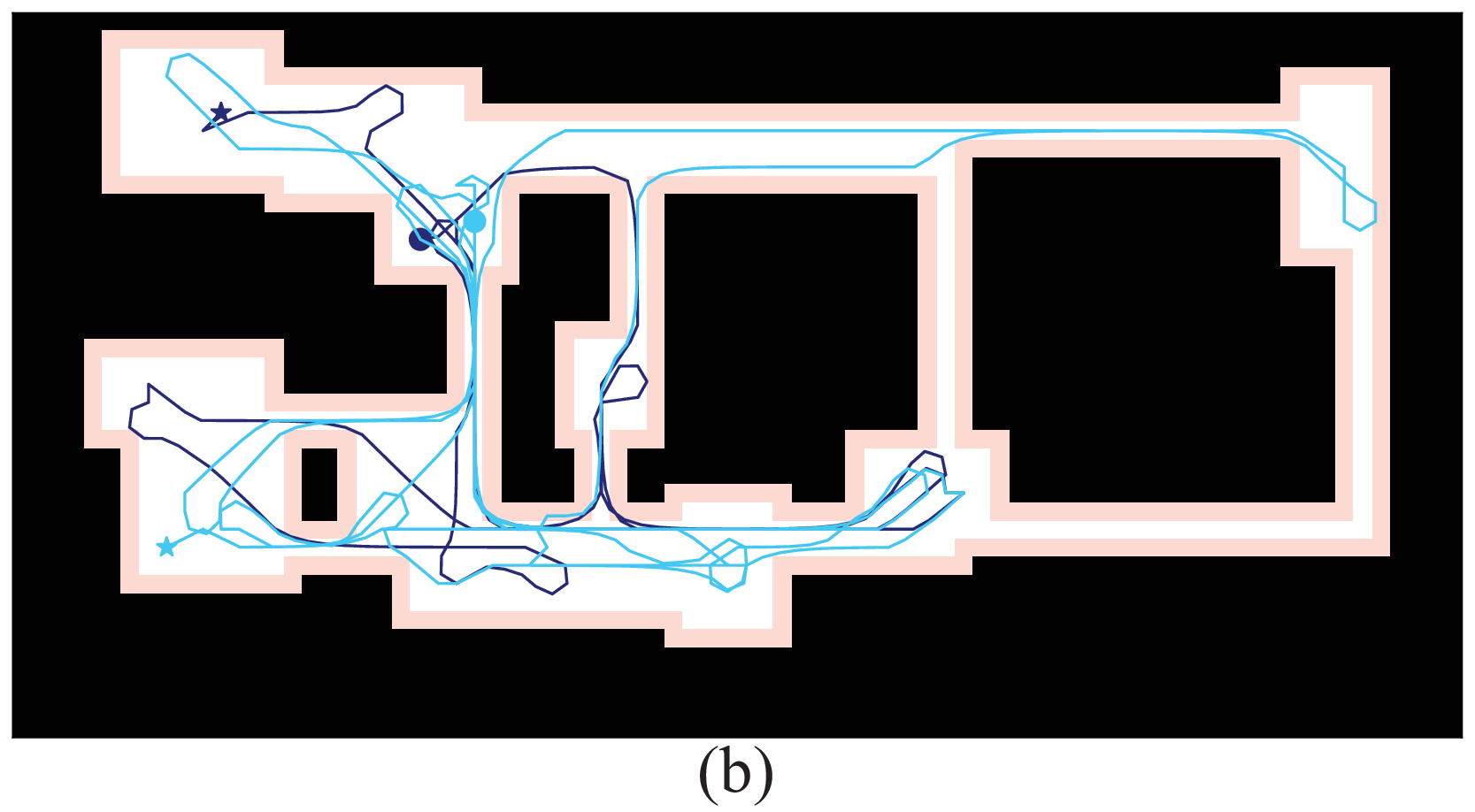}
  \label{fig:sim_b}
}
\subfloat{
  \includegraphics[width=0.3\textwidth]{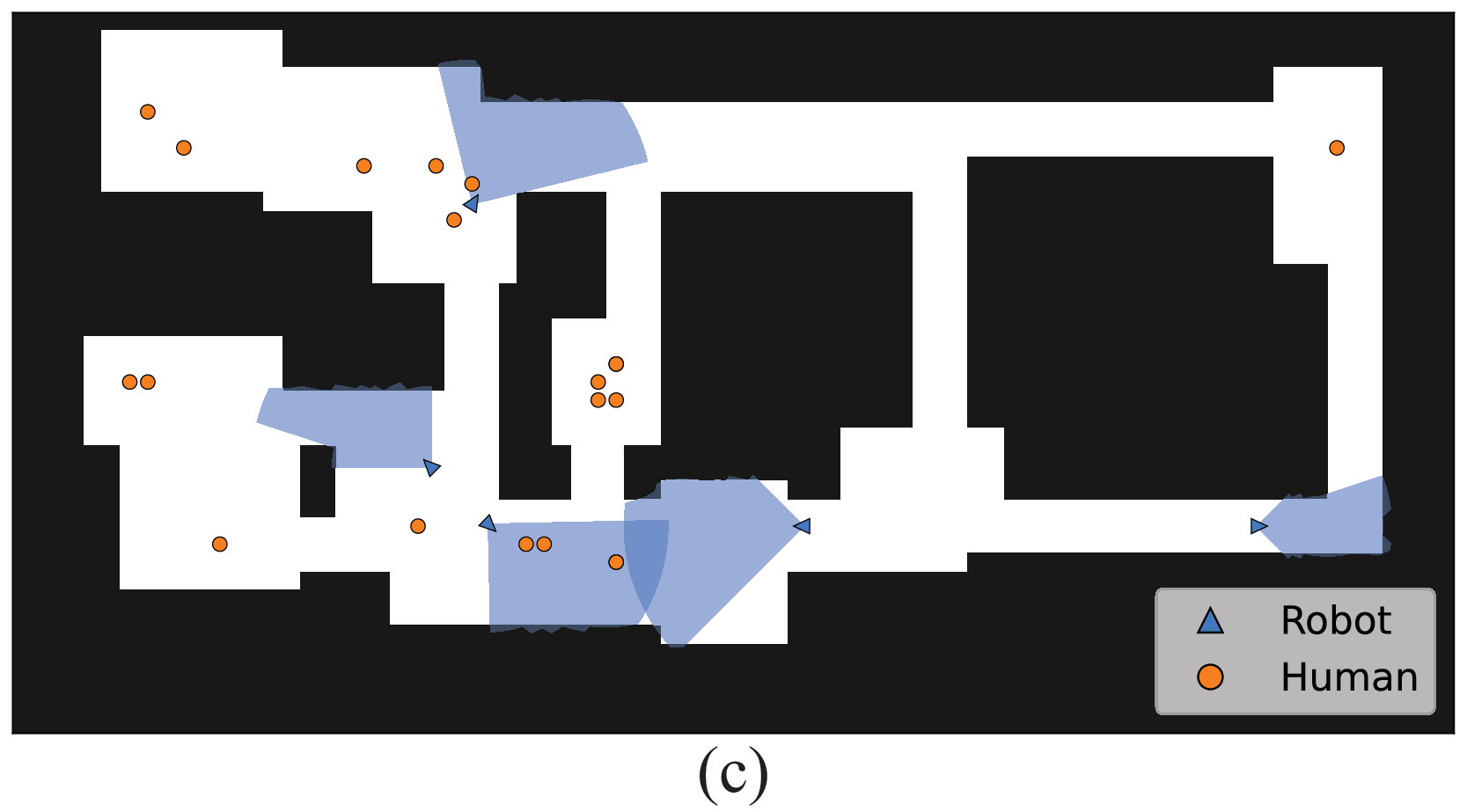}
  \label{fig:sim_c}
}
\caption{\textbf{An illustration of environment.} (a) A generated map showing the room and corridor structure, as well as distinct colored zones. (b) Synthetically generated human trajectories by a random hierarchical planner, avoiding the shaded buffer region to ensure safe paths. (c) An example of the simulation with robots and humans; the shaded blue regions represent the visible field of view (FoV) for each robot, as computed by the visibility function $f_\text{vis}$.}
\label{fig:exp3_sen_zone}
\vspace{-10pt}
\end{figure*}

\subsection{Multi-Robot Navigation and Coverage}
Multi-robot navigation and coverage have been studied under a variety of task objectives that emphasize coordinated motion and spatial coverage. 
One line of work studies multi-robot coverage path planning (MCPP), which coordinates robot paths to guarantee complete coverage of the environment while optimizing a makespan \cite{tang2021mstc}. Some extensions add objectives such as regional priorities \cite{lee2026priority} or turn minimization \cite{vandermeulen2019turn}. Persistent monitoring (PM) moves beyond single-pass coverage, designing periodic trajectories to ensure bounded revisit times or sustained observation frequency over space \cite{smith2011persistent, lan2013planning}. Other task settings include search-and-rescue, which emphasizes efficient search over large or uncertain regions \cite{wu2024autonomous}, perimeter defense or surveillance with spatiotemporal guarding responsibilities \cite{kim2025human, velhal2022decentralized}, and encirclement tasks that focus on coordinated positioning around a target \cite{gou2025policy}.
Exploration methods instead focus on frontier expansion in the observed workspace \cite{chiun2025marvel}. While these approaches address diverse navigation objectives, they largely optimize coverage or task execution metrics. In contrast, our work focuses on cooperative informative sensing, explicitly optimizing robot motion to improve human-centric monitoring accuracy under partial observability.

\section{Problem Formulation}\label{sec:formulation}

We formulate the multi-robot cooperative informative sensing problem as a Decentralized Partially Observable Markov Decision Process (Dec-POMDP, \cite{bernstein2002complexity}). The objective is to learn cooperative sensing policies for a team of robots to accurately predict key measures associated with dynamic human agents in an indoor environment, such as occupancy or flow patterns. 
Each robot agent is equipped with a 2D LiDAR sensor for obstacle detection, including indoor structures, and a front-facing camera for human detection.
The process is defined by the tuple $\langle \mathcal{I}, \mathcal{S}, \boldsymbol{\mathcal{O}}, \boldsymbol{\mathcal{A}}, \mathcal{R}, \mathcal{T}, \gamma \rangle$.

\subsubsection{Agent}
$\mathcal{I}$ denotes the set of $n$ decision-making robot agents, written as $\mathcal{I}\equiv\mathcal{I}_R = \{1, \dots, n\}$. The environment also contains $m$ non-controllable human agents $\mathcal{I}_H = \{1, \dots, m\}$, who follow their own distinct paths.

\subsubsection{State}
A state $s_t \in \mathcal{S}$ represents the complete configuration at timestep $t$, including the indoor environment layout and all agent states.
The indoor layout $\Omega \subset \mathbb{R}^2$ represents the static and collision-free workspace.
The kinematic state for each agent $i \in \mathcal{I}_R \cup \mathcal{I}_H$ consists of its pose $(p_t^i, \theta_t^i)$ and linear speed $v_t^i \in [0, v_{\max}^i]$. Here, $p_t^i \in \Omega$ is the $x\text{-}y$ position and $\theta_t^i \in [0, 2\pi)$ is the orientation. The state for each robot agent additionally includes the measurement received by the 2D LiDAR sensor for that robot, $d_t^i \in \mathbb{R}^L$, which is a vector of distances representing the $L$ LiDAR beams. 
Therefore, the overall global state is defined as:
\begin{equation} \label{eq:state_definition}
   s_t = (\Omega, \{ (p_t^i, \theta_t^i, v_t^i, d_t^i) \}_{i \in \mathcal{I}_{R}}, \{ (p_t^j, \theta_t^j, v_t^j) \}_{j \in \mathcal{I}_{H}}).
\end{equation}

\subsubsection{Observation}  $\boldsymbol{\mathcal{O}}=\times_{i \in \mathcal{I}_R} \mathcal{O}_i$ is the set of joint observations of all robot agents. Each robot agent $i \in \mathcal{I}_R$  receives a local observation $o_t^i\in \mathcal{O}^i$, which includes the precise pose of agent $i$, the agent's LiDAR measurements, and noisy measurements of currently visible human agents. Visibility of human $j\in\mathcal{I}_H$ from robot $i\in\mathcal{I}_R$ within the layout $\Omega$ is determined by the visibility function:
\begin{equation}
    f_{\text{vis}}(p_t^i,\theta_t^i,p_t^j, \Omega) \in \{0, 1\}.
\end{equation}
The visibility function returns $1$ if human $j$ is visible to robot $i$, accounting for sensor constraints and environmental occlusions based on the robot pose $(p_t^i, \theta_t^i)$ and human position $p_t^j$.
Let $\mathcal{I}_{H,t}^i$ be the set of indices of humans visible to robot $i$ at time $t$:
\begin{equation} \label{eq:visible_set}
   \mathcal{I}_{H,t}^i = \{ j \in \mathcal{I}_H \mid f_{\text{vis}}(p_t^i,\theta_t^i,p_t^j, \Omega) = 1 \}.
\end{equation}
Then, the local observation $o_t^i$ is defined as follows:
\begin{equation}\label{eq:observation}
    o_t^i = (\Omega, \{ (p_t^i, \theta_t^i, v_t^i, d_t^i) \}_{i \in \mathcal{I}_{R}}, \{ (\tilde{p}_t^j, \tilde{\theta}_t^j, \tilde{v}_t^j) \}_{j \in \mathcal{I}_{H,t}^i}),
\end{equation}
where $(\tilde{p}_t^j, \tilde{\theta}_t^j, \tilde{v}_t^j)$ is the noisy measurement of the true pose $(p_t^j, \theta_t^j)$ and velocity $v_t^j$ of visible human $j$, assumed drawn from a sensor noise model, e.g., $\mathcal{N}((p_t^j, \theta_t^j), \Sigma_{\text{sensor}})$.

\subsubsection{Action} $\boldsymbol{\mathcal{A}} = \times_{i \in \mathcal{I}_R} \mathcal{A}^i$ is the set of joint actions of all robot agents. The individual action $a_t^i \in \mathcal{A}_i$ for robot $i$ controls its desired linear speed $v_{\text{cmd}} \in \{ 0.0, 1.0, 2.0 \} \text{ m/s}$ and steering command $\delta_{\text{cmd}} \in \{ -\frac{\pi}{8}, 0.0, +\frac{\pi}{8} \} \text{ rad/s}$. The selected action serves as input to the low-level controller.

\subsubsection{Reward}

Across tasks, we maintain a persistent team belief $b_t$ over human-related latent variables, which is updated using newly acquired observations at time $t$. Each task defines a deterministic estimator $\Phi(\cdot)$ that maps the current belief to a task-level measure $m_t = \Phi(b_t)$. Active improvement is quantified as the change in this measure after the belief update as follows:
\begin{equation}
r_t = \lVert m_t - m_{t-1} \rVert_1.
\label{eq:active_improvement_reward}
\end{equation}
This reward encourages behaviors that actively refine task-relevant estimates through informative observations, rather than optimizing a static state-based objective. As a concrete example, in the human persistent monitoring task, $\Phi$ is the identity mapping so that $m_t$ corresponds to the persistent belief state. Then, the reward becomes the total belief update magnitude. During training, the environment internally accesses ground-truth human states to update $b_t$, while the policy receives only local observations.

\subsubsection{Transition} Function $\mathcal{T}: \mathcal{S} \times \mathcal{A} \to P(\mathcal{S})$ determines the probability distribution over the next state $s_{t+1}$ given the current state $s_t$ and joint action $\mathbf{a}_t$. The transition model operates with a discrete time interval $\Delta t$.
Human agents $j \in \mathcal{I}_H$ follow their predefined distinct paths, determined by an external process independent of robot actions $\mathbf{a}_t$.
For each robot agent $i \in \mathcal{I}_R$, its state is updated based on its current state $(p_t^i, \theta_t^i, v_t^i)$ and the selected action $a_t^i = (v^i_{\text{cmd}}, \delta^i_{\text{cmd}})$. 
The low-level controller updates the actual linear speed towards the command $v^i_{\text{cmd}}$, resulting in $v_{t+1}^i$. The position and orientation are updated deterministically based on the current state and steering command $\delta^i_{\text{cmd}}$ as follows:
\begin{equation}
\left\{
\begin{aligned}
\theta_{t+1}^i &= \theta_t^i + \delta^i_{\text{cmd}} \cdot \Delta t, \\
p_{t+1}^{i,x} &= p_t^{i,x} + v_t^i \cos(\theta_t^i) \cdot \Delta t, \\
p_{t+1}^{i,y} &= p_t^{i,y} + v_t^i \sin(\theta_t^i) \cdot \Delta t,
\end{aligned}
\right.
\end{equation}
where $p_t^i=(p_t^{i,x}, p_t^{i,y})$. The overall next state $s_{t+1}$ combines the static layout $\Omega$ and the updated states of all agents.

The objective is to find an optimal joint policy $\pi^* = \{\pi_i^*\}_{i \in \mathcal{I}_R}$ that maximizes the expected discounted cumulative reward, where $\gamma \in [0,1)$ denotes the discount factor:
\begin{equation}\label{eq:objective_function}
\pi^* = \underset{\pi}{\mathrm{argmax}} \; \mathbb{E} \left[ \sum_{t=0}^{\infty} \gamma^t r_t \right].
\end{equation}

\section{Learning Cooperative Sensing Policies}
\begin{figure*}[t!]
    \centering
    \includegraphics[width=0.9\textwidth]{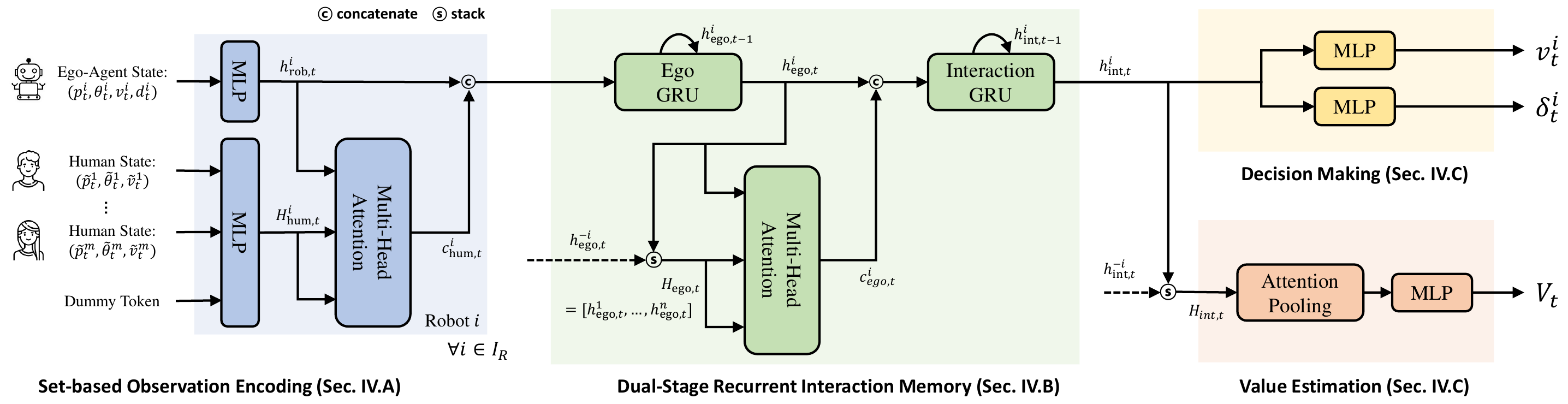}
    \vspace{-7pt}
    \caption{\textbf{Architecture of the proposed network.} 
(Left) Set-based observation encoding produces per-robot features via permutation-invariant attention. 
(Middle) A dual-stage recurrent interaction memory combines an ego GRU with attention-based inter-robot communication. 
(Right) The resulting features are used for per-robot decision making and centralized value estimation. The architecture is illustrated from the perspective of a robot $i$.}\label{fig:network}
\vspace{-10pt}
\end{figure*}

This section describes the proposed MARL framework for the problem formulated in \Cref{sec:formulation}. We first present the overall network architecture used in our actor-critic formulation, including the set-based observation encoding, dual-stage recurrent interaction memory, decision making, and value estimation modules, as shown in \Cref{fig:network}. We then detail the training procedure based on multi-agent proximal policy optimization (MAPPO, \cite{yu2022surprising}), outlining the optimization of the shared policy and centralized critic.

\subsection{Set-Based Observation Encoding}
Each robot $i\in\mathcal{I}_R$ receives a local observation $o_t^i$ at every timestep $t$, as defined in \Cref{eq:observation}, consisting of its ego state $(p_t^i, \theta_t^i, v_t^i, d_t^i)$ and a variable-sized set of visible human states $\{(\tilde p_t^j, \tilde \theta_t^j, \tilde v_t^j) \}_{j\in\mathcal{I}_{H,t}^i}$. 
Due to the varying number and ordering of visible humans across time and robots, human observations are modeled as an unordered set and encoded using a permutation-invariant architecture \cite{lee2019set}.
The ego state is embedded by an ego multilayer perceptron (MLP) into $h_{\mathrm{rob},t}^i\in\mathbb{R}^{d}$, and each human state is embedded by a shared MLP into token features of the same dimension. The tokens are then stacked into $H_{\mathrm{hum},t}^i \in \mathbb{R}^{N_t^i \times d}$ with $N_t^i = |\mathcal{I}_{H,t}^i| + 1$, where the additional learnable dummy token ensures a valid output even when no human is observed.
We aggregate human tokens via cross-attention using a multi-head attention ($\mathrm{MHA}$) network with the ego embedding as the query and the human tokens as keys and values, resulting in a context vector $c_{\mathrm{hum},t}^i\in\mathbb{R}^{d}$ as follows:
\begin{equation}
\label{eq:set_encoder_cross_attn}
c_{\mathrm{hum},t}^i
=
\mathrm{MHA}\!\left(
h_{\mathrm{rob},t}^i,\;
H_{\mathrm{hum},t}^i,\;
H_{\mathrm{hum},t}^i
\right).
\end{equation}
This design preserves permutation invariance and supports variable-sized sets via attention masking, without requiring persistent human identities or explicit tracking across timesteps. The concatenated representation $h_{\mathrm{rob},t}^i \,\Vert\, c_{\mathrm{hum},t}^i$ is then passed to the recurrent memory module.

\subsection{Dual-Stage Recurrent Interaction Memory}
To capture temporal dependencies while enabling inter-robot coordination, we maintain a dual-stage recurrent memory composed of two gated recurrent unit (GRU) components: an ego GRU ($\mathrm{GRU}_{\mathrm{ego}}$) followed by an interaction GRU ($\mathrm{GRU}_{\mathrm{int}}$). The ego GRU updates a private state from the concatenated per-robot representation as follows:
\begin{equation}
\label{eq:ego_gru}
h_{\mathrm{ego},t}^i
=
\mathrm{GRU}_{\mathrm{ego}}\!\left(h_{\mathrm{ego},t-1}^i,\; h_{\mathrm{rob},t}^i \,\Vert\, c_{\mathrm{hum},t}^i\right).
\end{equation}
Robots then exchange information by attending over the collection of ego hidden states. Let $H_{\mathrm{ego},t} = [h_{\mathrm{ego},t}^1;\dots;h_{\mathrm{ego},t}^{n}] \in \mathbb{R}^{n\times d}$ denote the stacked ego states. For each robot $i$, we compute a coordination context vector via multi-head attention with $h_{\mathrm{ego},t}^i$ as the query and the stacked ego states as keys and values:
\begin{equation}
\label{eq:inter_robot_mha}
c_{\mathrm{ego},t}^i
=
\mathrm{MHA}\!\left(
h_{\mathrm{ego},t}^i,\;
H_{\mathrm{ego},t},\;
H_{\mathrm{ego},t}
\right).
\end{equation}
Finally, the interaction GRU integrates this coordination signal over time to form an interaction-aware memory:
\begin{equation}
\label{eq:int_gru}
h_{\mathrm{int},t}^i
=
\mathrm{GRU}_{\mathrm{int}}\!\left(
h_{\mathrm{int},t-1}^i,\;
h_{\mathrm{ego},t}^i \,\Vert\, c_{\mathrm{ego},t}^i
\right).
\end{equation}
This factorization preserves a robot-specific temporal state while incorporating coordination information through attention, enabling scalable multi-robot interaction without requiring explicit pairwise tracking or fixed ordering.

\subsection{Decision Making and Value Estimation}
The interaction-aware representation $h_{\mathrm{int},t}^i$ is used for per-robot decision making, while value estimation is performed in a centralized manner.
The policy maps $h_{\mathrm{int},t}^i$ to independent categorical distributions over the linear and angular velocity action sets defined in \Cref{sec:formulation}.
For a centralized value estimation, the interaction states are stacked as $H_{\mathrm{int},t} = [h_{\mathrm{int},t}^1;\dots;h_{\mathrm{int},t}^{n}] \in \mathbb{R}^{n\times d}$, and a learnable query vector $q_V \in \mathbb{R}^{d}$ attends to this set via multi-head attention to obtain a pooled global representation:
\begin{equation}
z_t = \mathrm{MHA}\!\left(q_V,\; H_{\mathrm{int},t},\; H_{\mathrm{int},t}\right),
\end{equation}
which is subsequently mapped through an MLP to produce a scalar value estimate $V_t$.

\subsection{Multi-Agent Reinforcement Learning}

We train the proposed architecture using MAPPO \cite{yu2022surprising} under centralized training with a shared policy $\pi_\theta$ across all robots and a separately parameterized centralized critic $V_\phi$. While each robot executes actions based on its own interaction-aware representation, value estimation is performed centrally during training. Policy updates follow the clipped PPO objective with generalized advantage estimation (GAE) for variance reduction. Let $\hat{A}_t^i$ denote the advantage estimate for robot $i$ and $r_t^i(\theta)=\frac{\pi_\theta(a_t^i \mid o_t^i)}{\pi_{\theta_{\text{old}}}(a_t^i \mid o_t^i)}$ the probability ratio. The objective function is defined as follows:
\begin{equation}
\begin{aligned}
\mathcal{L}(\theta, \phi)
=
\mathbb{E}_t
\Big[
&\min \big(
r_t^i(\theta)\hat{A}_t^i,\,
\mathrm{clip}(r_t^i(\theta), 1-\epsilon, 1+\epsilon)\hat{A}_t^i
\big) \\
&- c_v \mathcal{L}_V(\phi)
+ c_e \mathcal{H}(\pi_\theta)
\Big].
\end{aligned}
\end{equation}
where $\mathcal{L}_V(\phi)$ is the value regression loss and $\mathcal{H}(\pi_\theta)$ denotes an entropy regularizer. 
To stabilize centralized value estimation, the critic is trained with access to ground-truth human states.
Visibility indicators are augmented as one-hot encodings, and all humans are assumed observable when computing value targets. This augmentation is used only during training and does not affect execution, where each robot relies solely on its local observation.

\section{Experiments}

\subsection{Experimental Setup}
\label{subsec:setup}

\begin{figure*}[t!]
    \centering
    \includegraphics[width=.9\textwidth]{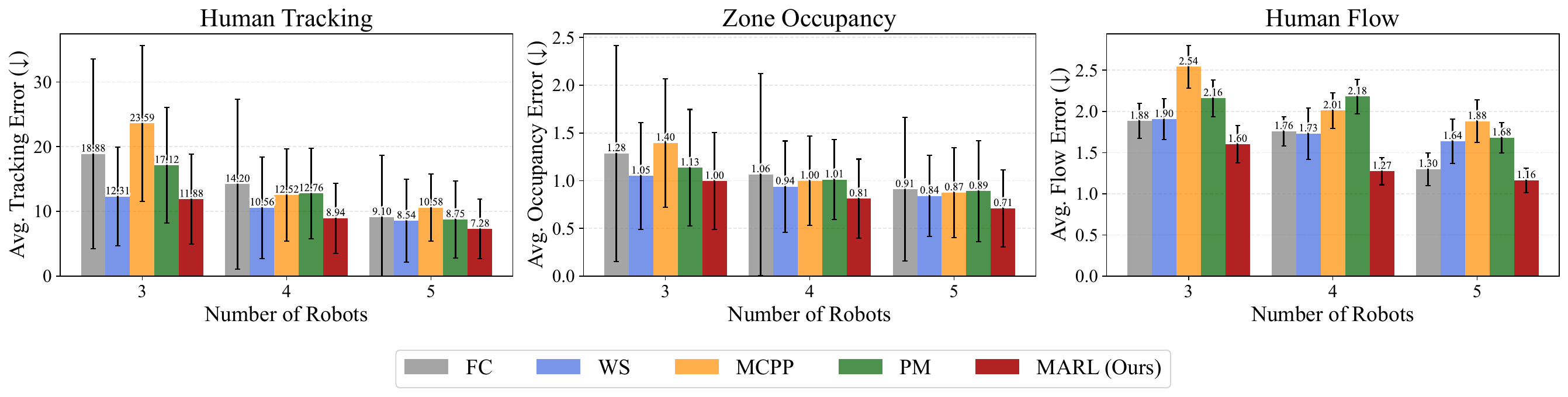}
    \vspace{-10pt}
    \caption{\textbf{Cooperative informative sensing performance comparison.} Average errors in human tracking (left), zone occupancy (middle), and human flow (right) monitoring tasks across different numbers of robots (3–5). Lower is better. Error bars indicate standard deviation.}
    \label{fig:main_results}
    \vspace{-7pt}
\end{figure*}

\begin{table}[t!]
\centering
\caption{Experimental Setup and Implementation Details}
\vspace{-5pt}
\resizebox{.9\columnwidth}{!}{%
\begin{tabular}{ll}
\toprule
\multicolumn{2}{l}{\textbf{Environment \& Sensing}} \\
\midrule
Map / \#Rooms / \#Zones & $80\times40$\,m / 12 / 7 \\
Time step / horizon & $\Delta t=1$\,s, $T=500$ \\
\#Humans / \#Robots & $M=20$, $N=5$ \\
Observation noise & $\sigma_p=0.2$\,m, $\sigma_\theta=0.1$\,rad \\
Action space & $v \in \{0,1,2\}$\,m/s (max 2.0) \\
            & $\omega \in \{-\pi/8,0,\pi/8\}$\,rad/s \\
Tracking sensor & range 10\,m, FoV $90^\circ$, \\
               & occlusion via ray sampling ($K=5$) \\
LiDAR & range 10\,m, FoV $360^\circ$, 16 rays \\
\midrule
\multicolumn{2}{l}{\textbf{Network Architecture \& MAPPO Training}} \\
\midrule
Network hidden dim & 64\\
Learning rate & $3\times10^{-4}$ (Actor / Critic)\\
MAPPO / Optimization & clip 0.2, max grad norm 0.5, value coef 0.5 \\
Discount / GAE & $\gamma=0.99$, $\lambda=0.95$ \\
Entropy regularization & 0.01 (linear speed), 0.001 (rotation) \\
Rollout / Updates & 250 parallel environments, 1000 steps per update, \\
                 & 20 minibatches, 20 epochs, 50 chunk length \\
                 & 50M timesteps \\
\bottomrule
\end{tabular}
}
\label{tab:setup_details}
\vspace{-10pt}
\end{table}

To evaluate cooperative informative sensing policies, we develop a simplified 2D simulation environment. We generate indoor environments with room and corridor structures \cite{merrell2010computer}, avoiding unrealistic cave-like topologies. We sequentially carve randomly sized, overlapping rooms and connect their centroids with corridors to ensure the entire map $\Omega$ is connected. Physically overlapping rooms are grouped into distinct zones as shown in \Cref{fig:sim_a}. These environments are then populated with synthetic human paths generated using a hierarchical model. A high-level planner selects goal destinations via a Markov chain \cite{song2010limits} across the rooms and samples realistic dwell times from a log-normal distribution \cite{ashbrook2003using}. A low-level executor then generates plausible, collision-free motion by following an A$^*$ path \cite{hart1968formal} with a pure pursuit controller, as shown in \Cref{fig:sim_b}.
Robot partial observations are implemented via ray sampling, which approximates occlusions by checking $K$ parallel points to avoid slow sequential ray-tracing. An example of the whole environment is shown in \Cref{fig:sim_c}. The complete experimental configuration, including environment parameters and MARL training hyperparameters, is summarized in \Cref{tab:setup_details}.

\subsection{Description of Monitoring Tasks and Baselines}\label{sec:exp_desc}
We evaluate our methods on three indoor monitoring tasks that capture complementary aspects of indoor human activity.  Each task emphasizes a different sensing granularity: tracking individual humans, estimating zone-level occupancy, and capturing flow between zones.
As defined in \Cref{sec:formulation}, the team maintains a belief and derives a task-level estimate $m_t$ under partial observability. 
The magnitude of belief updates defines the training reward, and the task-specific estimators are specified below. Let $Z$ denote the number of semantic zones presented in \Cref{fig:sim_a}.

\begin{itemize}[leftmargin=*]
    \item \textbf{Human Persistent Monitoring}: 
    We maintain a persistent per-human position belief $\hat{p}^j_t \in \mathbb{R}^2$ for each human $j$. If any robot observes human $j$ at time $t$, $\hat{p}^j_t$ is updated to the current position and otherwise remains unchanged. The task-level estimate is $m_t = [\hat{p}^1_t;\ldots;\hat{p}^M_t]$.

    \item \textbf{Zone Occupancy Monitoring}: 
    We maintain zone-level occupancy estimates derived from the persistent position belief $\hat{p}^j_t$. 
    Each estimate is mapped to a zone, and the task-level estimate 
    $m_t \in \mathbb{R}^{Z}$ is defined as the zone-wise occupancy count vector,
    $m_t[z] = \sum_{j=1}^{M}\mathbb{I}[\hat{p}^j_t \in \text{zone } z]$.

    \item \textbf{Human Flow Monitoring}: We maintain a per-human zone-transition belief $(\hat{z}^{j,\text{prev}}_t,\hat{z}^{j,\text{cur}}_t,\hat{\tau}^j_t)$, updated only when an observed human changes zones. Upon a transition, we record the previous and current zones and the transition time. The task-level estimate $m_t \in \mathbb{R}^{Z \times Z}$ is the directed flow count over all zone pairs, defined as $m_t[z_1,z_2] = \sum_{j=1}^{M} \mathbb{I}[\hat{z}^{j,\text{prev}}_t = z_1 \land \hat{z}^{j,\text{cur}}_t = z_2]$.
\end{itemize}
For all tasks, the per-step reward component is clipped to a fixed range to prevent extreme updates and stabilize training.

We compare our MARL-based active observation policy with baseline planners, covering both classical and learning-free approaches to evaluate the impact of learned coordination on information gathering efficiency.
\begin{itemize}[leftmargin=*]
    \item \textbf{FC} (Fixed Cameras) uses static sensors placed to cover the largest visible zones without overlap, serving as a reference for non-mobile sensing performance. Sensors are placed sequentially using a greedy algorithm to maximize the area not yet covered by previously placed static sensors.
    
    \item \textbf{WS} (Waypoint Sampling) imitates human high-level planning by randomly selecting goal points within and across rooms and navigating between them sequentially, resulting in spatially diverse and natural exploration behavior.
    
    \item \textbf{MCPP} (Multi-Robot Coverage Path Planning) uses a spanning tree coverage strategy~\cite{tang2021mstc} to achieve efficient coverage of the environment. The tree is partitioned among robots, with path endpoints connected to form closed loops that are repeatedly traversed while minimizing makespan.
     
    \item \textbf{PM} (Persistent Monitoring) uses the solution of a linear program to control robot velocities along closed paths, guaranteeing a minimum bounded frequency that a space is observed by the robots \cite{smith2011persistent}.
\end{itemize}

\subsection{Cooperative Informative Sensing Performance}

We evaluate the proposed cooperative informative sensing algorithm by comparing it with baseline methods across the three monitoring tasks described in \Cref{sec:exp_desc}. To assess performance under different sensing resources, we vary the number of robots from 3 to 5 while keeping the environment and human dynamics fixed. The proposed MARL policy is trained with five robots and evaluated in a zero-shot manner for different team sizes (e.g., $n=3,4$), and the same trained model is used for all subsequent experiments. \Cref{fig:main_results} shows the comparative performance across methods and team sizes, with results averaged over 250 scenarios.

Overall, the proposed MARL policy achieves the lowest average error across all tasks and evaluated team sizes, including zero-shot settings. As the number of robots increases, the errors generally decrease for all methods and the performance gaps across algorithms become smaller, indicating that additional sensing resources reduce overall difficulty. A notable observation is that FC exhibits substantially larger standard deviations in human tracking and zone occupancy, suggesting that its mean performance is more sensitive to scenario variations. 
PM exhibits a non-monotonic trend in human flow estimation, where the error slightly increases from $2.16$ at $n=3$ to $2.18$ at $n=4$, indicating that additional robots do not always lead to improved coordination.

\subsection{Robustness under Out-of-Distribution (OOD) Scenario}

\begin{figure}[t!]
    \centering
    \includegraphics[width=.85\columnwidth]{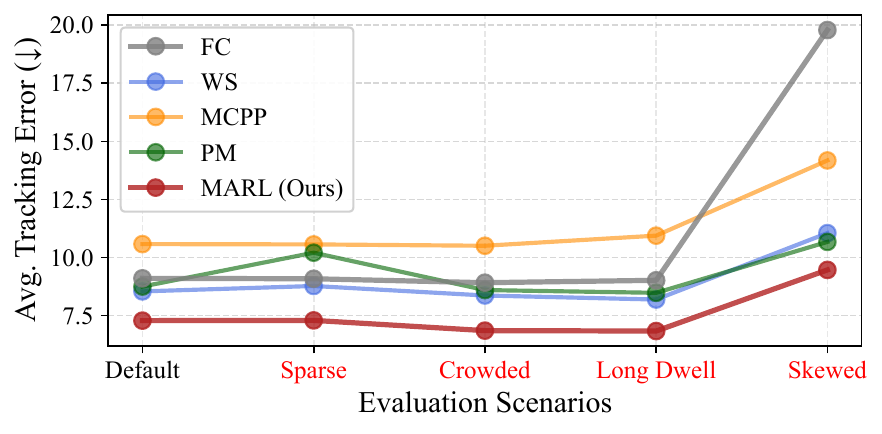}
    \vspace{-10pt}
    \caption{\textbf{Evaluation under OOD scenarios.} Average tracking error across the default setting and four OOD variants involving changes in population size and human movement distributions (OOD cases shown in red).}
    \vspace{-15pt}
    \label{fig:ood_results}
\end{figure}

We evaluate if the proposed MARL policy maintains strong monitoring performance under distribution shifts that were not observed during training. To this end, we construct four OOD scenarios that perturb either the number of humans or their behavioral dynamics while keeping the sensing and environment configurations unchanged. 
Specifically, we vary the human population from the default 20 to 10 (Sparse) and 30 (Crowded), increase dwell times to induce more persistent stationary behavior (Long Dwell), and amplify the transition probability toward a spatially isolated right-side zone (zone 3 in \Cref{fig:sim_a}) by $2$ times (Skewed).
As shown in \Cref{fig:ood_results}, most methods show only minor variations under Sparse, Crowded, and Long Dwell, indicating that moderate shifts in population size or dwell-time persistence do not significantly affect monitoring performance.
In contrast, performance decreases under the Skewed scenario for all methods. It suggests that visitation concentrated in spatially isolated regions is more difficult to monitor. FC shows the largest drop, suggesting limited adaptability when such regions become more frequently used. In comparison, our MARL policy achieves the lowest tracking error across all scenarios, including Skewed, demonstrating consistent performance under distribution shifts.

\subsection{Scalability and Marginal Utility Analysis}

\begin{figure}[t!]
\centering
\includegraphics[width=1.\columnwidth]{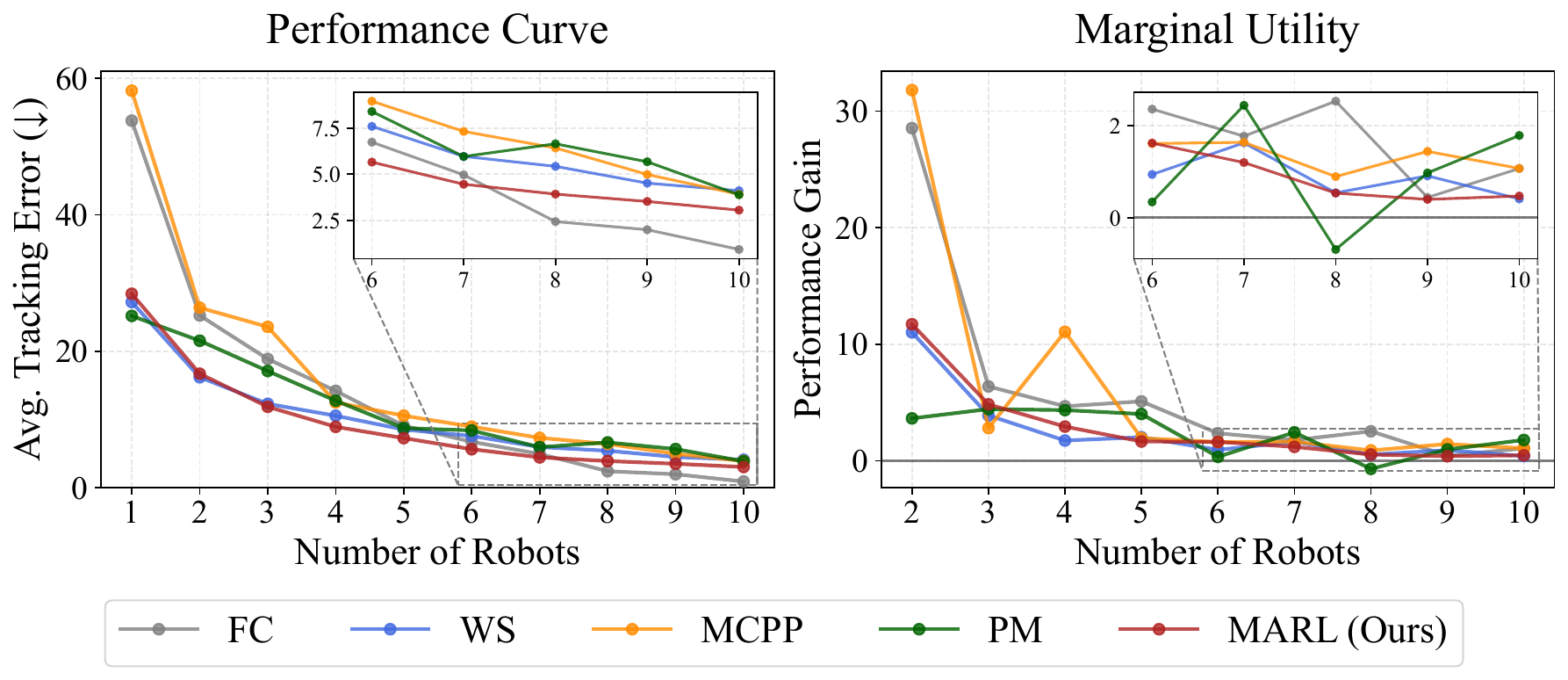}
\vspace{-15pt}
\caption{\textbf{Scalability and marginal utility analysis.} Tracking error (left) and marginal performance gain (right) with increasing number of robots.}
\label{fig:sens_results}
\vspace{-10pt}
\end{figure}

To evaluate scalability and the benefit of additional robots, we vary the number of robots from 1 to 10 while keeping the environment and human dynamics fixed. \Cref{fig:sens_results} shows the average tracking error (left) and the marginal utility (right) as the number of robots increases. We define the marginal utility at team size $n$ as the reduction in tracking error when increasing the number of robots from $n-1$ to $n$ as follows:
\begin{equation}
\Delta(n) = E(n-1) - E(n),
\end{equation}
where $E(n)$ denotes the average tracking error with $n$ robots.
Overall, performance improves as more robots are deployed, indicating that additional sensing resources enhance spatial coverage, while the marginal utility quickly saturates as the team size grows.
In this map configuration, the marginal gain becomes small beyond approximately $5$ robots, showing diminishing returns as coverage redundancy increases.
FC achieves the lowest tracking error at larger team sizes (e.g., $n \geq 8$), reflecting the advantage of globally optimized fixed placements that minimize overlap.
In contrast, mobile sensing approaches exhibit less consistent scaling behavior, suggesting that coordination overhead limits the effective utilization of larger teams. Notably, PM shows a slight performance degradation at $n=8$, indicating instability at higher team sizes.
The MARL policy maintains consistently strong performance across all team sizes without significant degradation, demonstrating robust scalability.

\subsection{Cooperation with Existing Sensing Systems}

\begin{figure}[t!]
\centering
\includegraphics[width=.85\columnwidth]{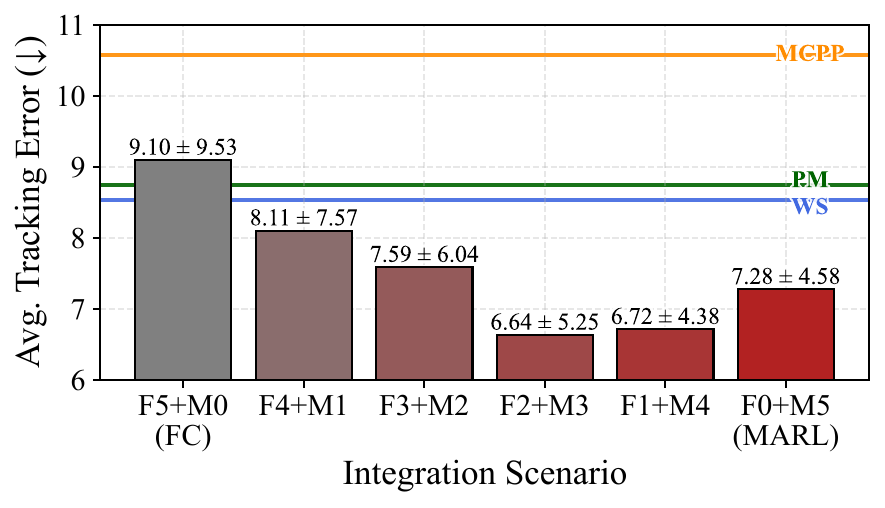}
\vspace{-10pt}
\caption{\textbf{Cooperation with an existing sensing system.} 
Average tracking error under hybrid integration of fixed cameras (F) and mobile robots controlled by MARL policy (M) with a total budget of five sensors. Horizontal lines indicate standalone baselines.}
\vspace{-5pt}
\label{fig:coop_results}
\end{figure}

We study cooperation with an existing fixed sensing system by integrating fixed cameras with MARL-controlled mobile robots under a fixed sensing budget. Concretely, we consider several hybrid integration scenarios with a total budget of five sensors, ranging from fixed-only ($F5{+}M0$) to mobile-only ($F0{+}M5$), including mixed configurations such as $F2{+}M3$ and $F1{+}M4$. We trained a MARL policy for each configuration. \Cref{fig:coop_results} summarizes the results, where the leftmost bar corresponds to FC ($F5{+}M0$) and the rightmost bar corresponds to MARL ($F0{+}M5$).

\begin{figure}[t!]
\centering
\subfloat{
  \includegraphics[width=.82\columnwidth]{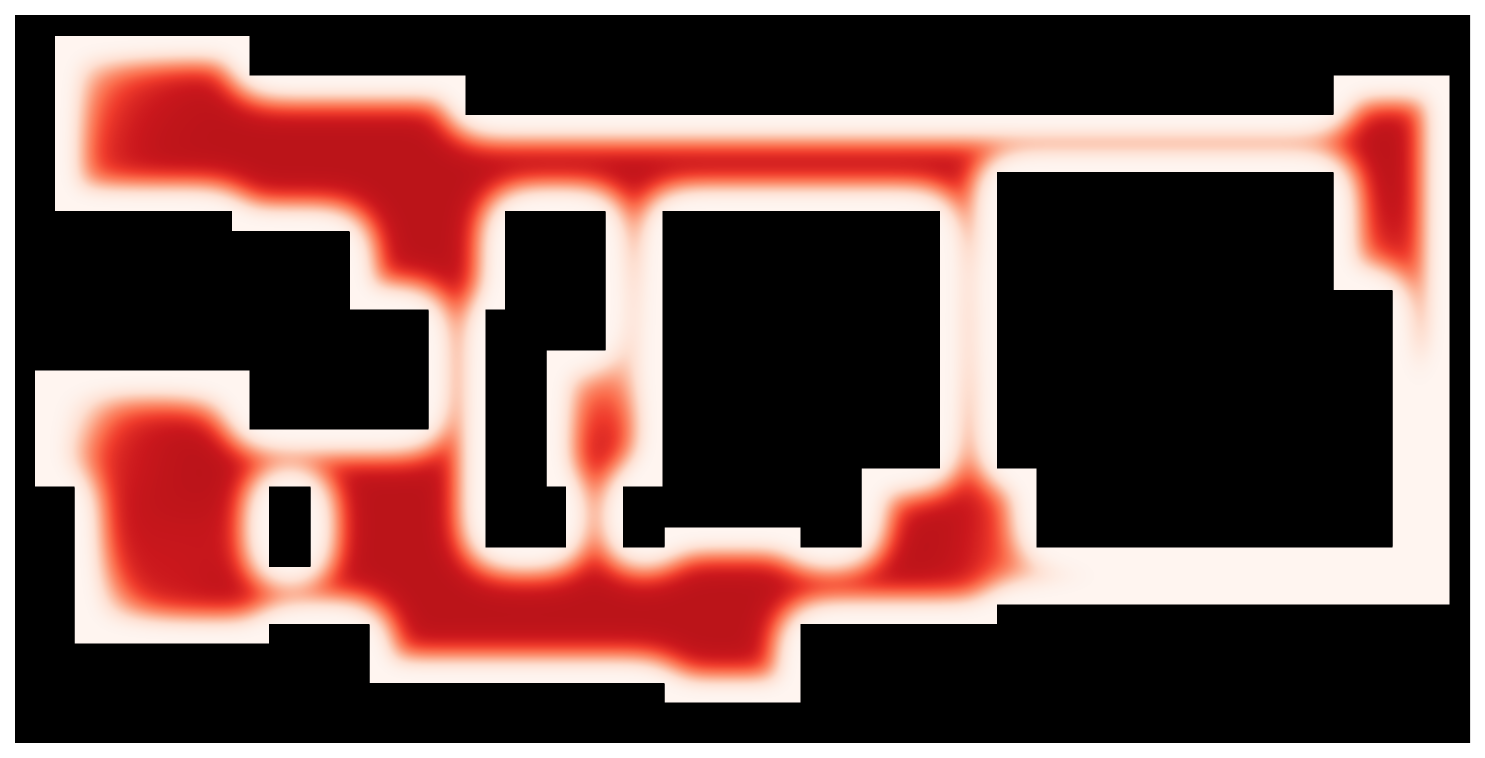}
  \label{fig:coop_results_heatmap_a}
}
\\
\vspace{-10pt}
\subfloat{
  \includegraphics[width=.82\columnwidth]{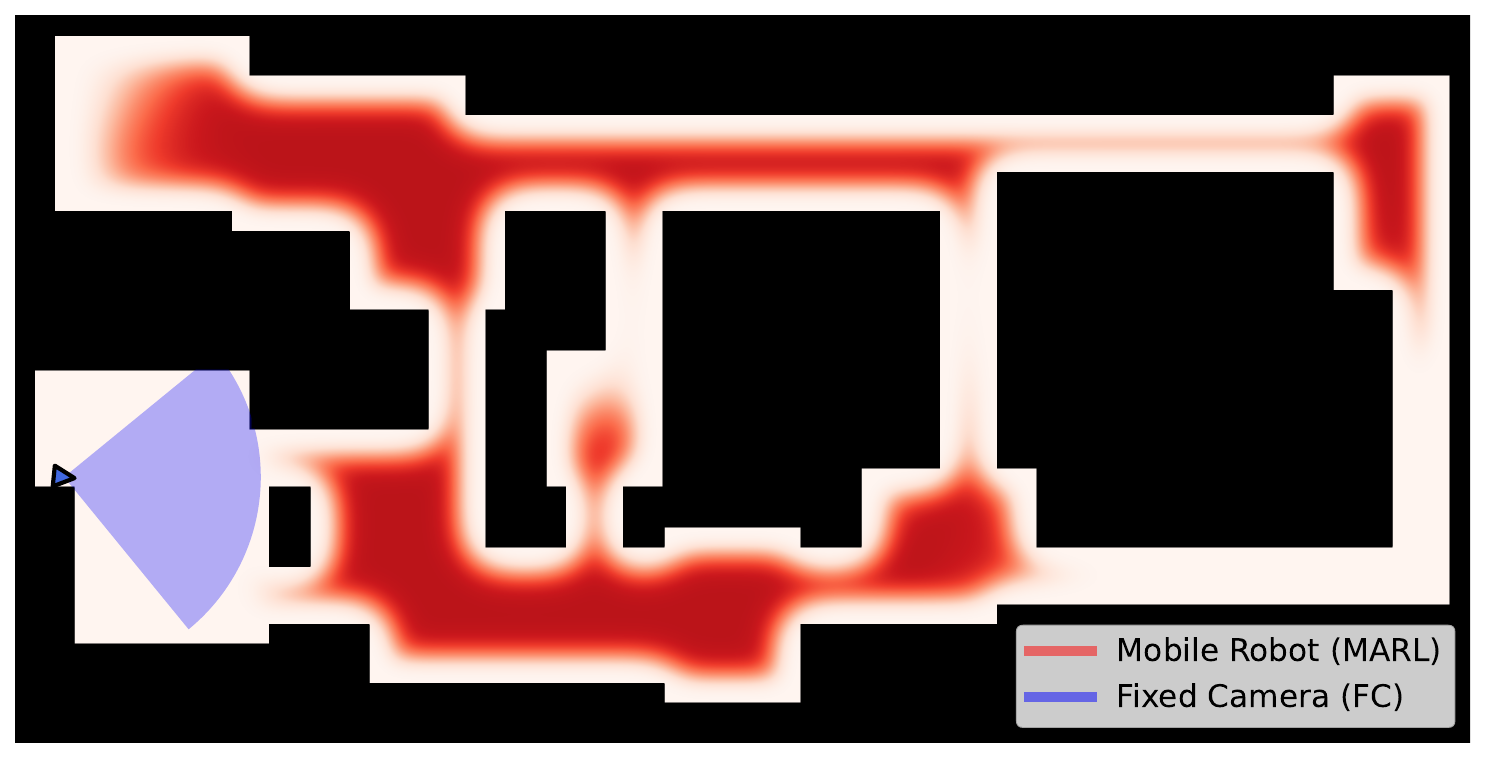}
  \label{fig:coop_results_heatmap_b}
}
\vspace{-5pt}
\caption{\textbf{Visibility heatmap under hybrid integration.}
Observed-region heatmaps for the mobile-only setting (F0+M5, top) and the hybrid setting (F1+M4, bottom). Red denotes regions observed by MARL-controlled mobile robots, and blue denotes the fixed camera FoV.}
\label{fig:coop_results_heatmap}
\vspace{-10pt}
\end{figure}

The results show that hybrid integration can outperform both fixed-only and mobile-only sensing. In particular, mixed configurations such as $F2{+}M3$ and $F1{+}M4$ achieve lower tracking error than using only fixed cameras or only mobile robots. 
Importantly, $F1{+}M4$ performs comparably to $F0{+}M5$ while incurring lower operational cost, since fixed infrastructure is typically cheaper to operate than an additional mobile robot.
This behavior is further illustrated in \Cref{fig:coop_results_heatmap}, where the observed-region heatmaps differ between $F0{+}M5$ and $F1{+}M4$. In the hybrid setting, mobile robots avoid regions already covered by the fixed camera and allocate their trajectories to complementary areas, showing adaptive coverage distribution. However, as the number of fixed cameras further increases, performance begins to degrade. When fixed sensors dominate the sensing budget, mobile robots have limited opportunity to contribute additional coverage, reducing the benefit of adaptive coordination. These results suggest that an appropriate balance between fixed and mobile sensing is necessary to maximize information gain, since learning an effective cooperative strategy is more difficult with a large number of robots.

\subsection{Impact of Fixed Camera Placement in the Hybrid Setting}

\begin{figure}[t!]
\centering
  \includegraphics[width=0.85\columnwidth]{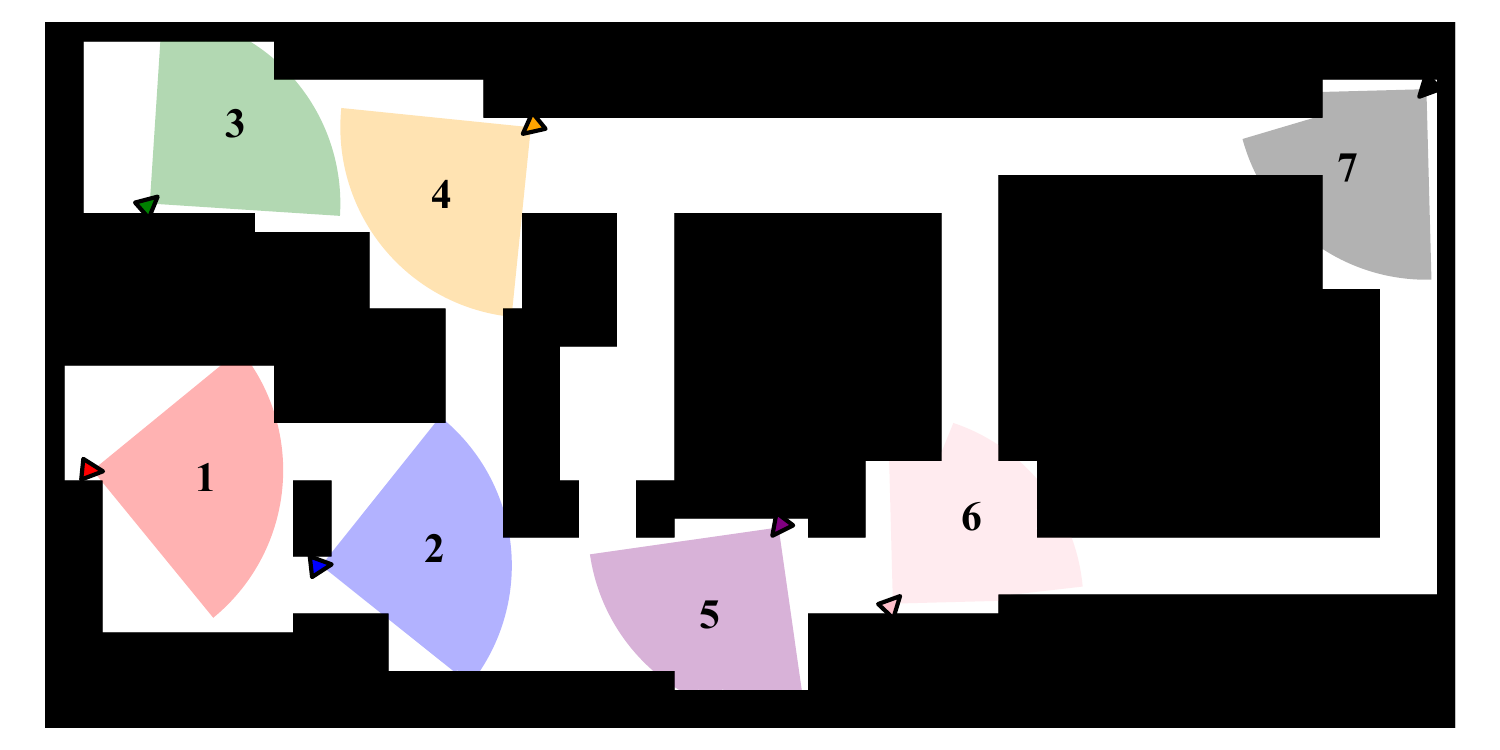}
  \vspace{-10pt}
\caption{\textbf{Fixed camera placement candidates (F1+M4).} We consider seven feasible camera placement IDs (1-7) in the same indoor layout. Each shaded region indicates the fixed camera FoV under that placement.}
\label{fig:fixed_camera_candidates}
\vspace{-5pt}
\end{figure}

To understand where to install a single FC in the hybrid setting (F1+M4), we evaluate seven feasible camera placements in the same indoor layout as shown in \Cref{fig:fixed_camera_candidates} and compare their tracking performance in \Cref{tab:position_performance}. Overall, placements located in outer regions (1, 3, and 7) of the map achieve the best mean performance. These placements cover areas that would otherwise require explicit robot visits. It reduces unnecessary traversal and allows the robots to concentrate on monitoring the remaining parts of the environment.
However, Placement 6 does not provide comparable improvement despite being away from the center, as it covers only a single traffic room, as shown in \Cref{fig:sim_a}.
In contrast, Placement 7 remains highly effective by covering a room at the end of a long corridor that is costly for robots to reach. Placement 7 also reduces the standard deviation, indicating improved robustness by consistently observing a frequently traversed yet hard-to-reach area.
On the other hand, central placements (2, 4, and 5) perform worse, as robots inevitably traverse the central region, making the FC largely redundant.
These results highlight that effective FC deployment should consider the indoor layout topology and robot traversal characteristics. In particular, placement plays a critical role in improving monitoring performance under a constrained sensing budget.

\begin{table}[t!]
\centering
\caption{Tracking Error Across Camera Position ID}
\vspace{-5pt}
\resizebox{1.0\columnwidth}{!}{%
\begin{tabular}{c||ccccccc}
\toprule
\textbf{Position Index} 
& 1 & 2 & 3 & 4 & 5 & 6 & 7 \\
\midrule
\textbf{Mean} 
& \textbf{6.72} & 7.79 & \textbf{6.83} & 7.33 & 8.46 & 7.57 & \textbf{6.76} \\
\textbf{Std} 
& 4.38 & 5.13 & 4.25 & 4.89 & 5.83 & 5.02 & \textbf{3.80} \\
\bottomrule
\end{tabular}
}
\label{tab:position_performance}
\vspace{-10pt}
\end{table}

\subsection{Reward Analysis}

\begin{figure}[t!]
\centering
  \includegraphics[width=0.8\columnwidth]{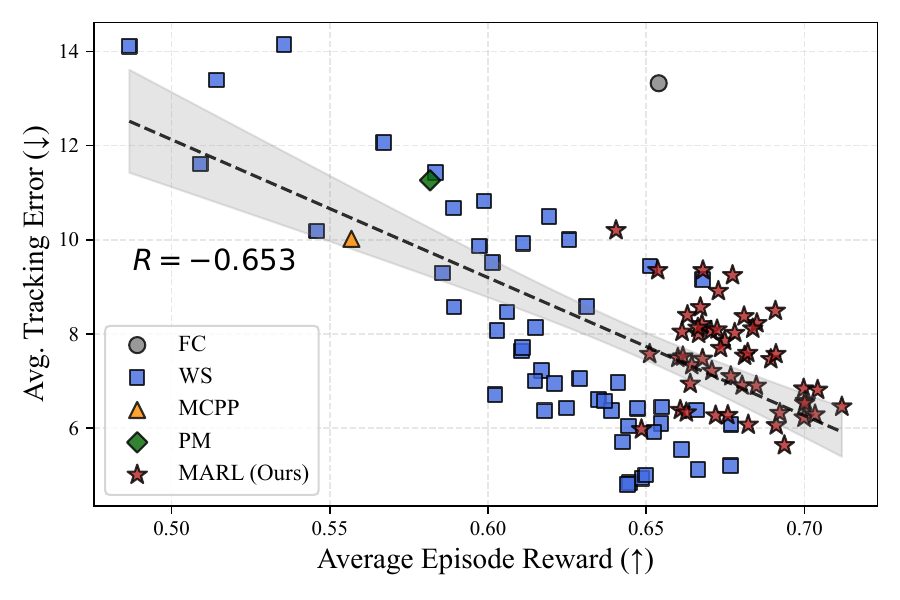}
  \vspace{-10pt}
\caption{\textbf{Correlation between reward and tracking error.} 
Each point represents a different robot behavior evaluated in the same scenario. 
The dashed line denotes linear regression and the shaded region indicates the 95\% confidence interval.}
\label{fig:reward_scatter}
\vspace{-12pt}
\end{figure}

To evaluate whether the proposed reward aligns with actual monitoring performance, we analyze the correlation between episode reward and tracking error under a fixed scenario while varying robot behaviors. For WS, behavioral diversity is introduced by sampling different human-following trajectories, and for MARL, multiple runs are conducted with different random seeds and stochastic policies. In contrast, FC, MCPP, and PM are deterministic and therefore correspond to single points in \Cref{fig:reward_scatter}. We observe a clear negative correlation between reward and tracking error ($R = -0.653$), with even stronger correlations for occupancy ($R = -0.80$) and flow ($R = -0.89$). Although the reward is not a perfect surrogate for the evaluation metric, the strong negative correlations indicate that increasing reward is consistently associated with improved monitoring performance.

\section{Conclusions and Discussions}

In this paper, we present a MARL-based cooperative informative sensing framework for dynamic indoor monitoring. Our approach optimizes observation accuracy across diverse human-centric tasks and highlights the importance of cost-effective hybrid integration with existing fixed infrastructure. In particular, FC placement must consider indoor layout topology, as redundant coverage can reduce monitoring gains.
While we evaluate in a 2D simulation to enable large-scale statistical validation, our architecture processes low-dimensional semantic states which guarantees that the cooperative policy remains structurally invariant when integrating real-world perception pipelines such as RGB-D object tracking. 
Future work will focus on deploying this modular framework in high-fidelity 3D simulators and real-world environments. We also aim to incorporate richer semantic information and complex human-centric constraints. In addition, we plan to leverage the reasoning capabilities of Large Language Models (LLMs) or Vision-Language Models (VLMs) to enhance agent explainability and generalizability.

\section*{Acknowledgement}
This research was supported by Center for Advanced Urban Systems (CAUS) of Korea Advanced Institute of Science and Technology (KAIST) funded by GS E\&C.
This work was also supported by the National Research Foundation of Korea (NRF) grant funded by the Korea government (MSIT) (No.RS-2025-00517342). This material is based upon work supported by the National Science Foundation (NSF) Graduate Research Fellowship Program under Grant No. DGE-2146755. This research is also partially supported by the Stanford’s Center for Sustainable Development and Global Competitiveness (SDGC).

\printbibliography

\end{document}